\documentclass{article} 
\usepackage[final]{colm2026_conference}

\usepackage[T1]{fontenc}
\usepackage[utf8]{inputenc}
\usepackage{latexsym}
\usepackage{amsmath}
\usepackage{graphicx}
\usepackage{enumitem}
\usepackage{booktabs}
\usepackage{microtype}
\usepackage{hyperref}
\usepackage{url}
\usepackage{lineno}

\definecolor{darkblue}{rgb}{0, 0, 0.5}
\hypersetup{colorlinks=true, citecolor=darkblue, linkcolor=darkblue, urlcolor=darkblue}

\title{Can LLMs Predict the Future? \\ A Brier Score Analysis of Prediction Markets}

\author{Yuanbo Li, Zekun Li\thanks{Equal contribution.}\, \& Xiaoyan Cong\footnotemark[1] \\
Brown University \\
Providence, RI 02912, USA \\
\texttt{yuanbo\_li@brown.edu}
}

\begin{document}

\ifcolmsubmission
\linenumbers
\fi

\maketitle

\begin{abstract}
We study whether model upgrades improve probability estimates for prediction-market questions.
Our \textbf{Resolved Market Forecasting (RMF)} benchmark contains $3{,}000$ resolved binary questions across nine domains, on which we evaluate six Claude and Qwen model variants using a question-only, zero-shot protocol.
We assess Brier scores relative to an empirical base-rate predictor, examine their Murphy decomposition, and compare models through paired differences, with results stratified by event category and timing relative to training cutoffs.
In the reported post-cutoff stratum, the four Claude models achieve Brier scores of $0.183$--$0.192$, improving on the base-rate reference by $0.024$--$0.033$.
Qwen 32B does not significantly outperform that reference, although its paired Brier is $0.024$ lower than that of the 7B checkpoint.
The evaluated Claude version and tier upgrades yield no significant improvement.
Within-model differences across event categories exceed the observed differences among Claude variants.
These results show why forecasting scores should be interpreted alongside simple probability baselines and question composition: under this protocol, newer versions or higher model tiers do not consistently produce more accurate probabilities.
\end{abstract}

\section{Introduction}

An event forecast communicates both an expected outcome and a degree of uncertainty.
Two models assigning probabilities of $0.55$ and $0.95$ to the same event make the same binary prediction, but express substantially different confidence.
Evaluating such outputs requires a score that retains the probability estimate and a reference that gives the resulting error a meaningful scale.
As LLMs are increasingly studied as probabilistic forecasters \citep{halawi2024approaching,schoenegger2024wisdom,karger2025forecastbench}, these choices become central to interpreting their performance.

\paragraph{Brier scores and base-rate comparisons.}
We use the Brier score, the mean squared difference between a forecast probability and the observed binary outcome \citep{brier1950verification}.
A low score alone does not establish how much information a model extracts from an individual question.
When outcomes are imbalanced, a constant forecast based on the marginal event frequency can already perform well.
We therefore compare model predictions with an empirical base-rate reference within each evaluation stratum.
This comparison asks whether the model improves on a forecast that ignores the content of every question.

Aggregate error also combines several properties of a forecast and its evaluation set.
The Murphy decomposition separates reliability, resolution, and uncertainty \citep{murphy1973new}: whether stated probabilities align with observed frequencies, whether forecasts distinguish groups with different outcome rates, and how balanced the outcomes are overall.
These components help interpret differences across event categories, where a change in raw Brier need not represent a change in calibration.
Our analysis uses this established framework to describe both the magnitude and the composition of forecasting error.

\paragraph{A shared question-only protocol.}
We construct \textbf{Resolved Market Forecasting (RMF)} from public Polymarket records, drawing a stratified sample of $3{,}000$ binary questions from more than $600{,}000$ resolved markets.
The records span nine domains and provide settlement outcomes, resolution dates, and category labels (Sec.~\ref{sec:setup}).
Each model receives the market question and returns a probability, with no retrieval or additional market context.
This fixed input protocol lets us examine what the evaluated models produce from internalized knowledge.
We will release the RMF data and evaluation code upon acceptance.

\paragraph{Comparing models, questions, and time periods.}
Our model selection supports two types of comparison: the 7B and 32B Qwen2.5 checkpoints differ in size, while the Claude Sonnet and Opus variants provide version and tier comparisons.
For each pair, we compute the Brier difference on questions answered by both models and report a bootstrap confidence interval.
Pairing keeps the question set fixed when assessing a model change; category breakdowns then show where aggregate comparisons conceal variation across topics.
The empirical base-rate reference and paired model comparisons answer complementary questions: whether a model adds predictive information, and whether a different model improves the score on the same questions.

The resolved records also permit retrospective comparisons across the temporal strata specified in Sec.~\ref{sec:setup}.
We examine how scores and their components vary with event timing relative to training cutoffs, alongside category composition, market salience, and agreement between models.
The model identifiers, cutoff assumptions, and sample definitions are provided in the setup and appendices.

The study contributes a question-only benchmark and an analysis that links aggregate Brier scores to baseline performance, paired model effects, and variation across evaluated events.
Section~\ref{sec:capacity} addresses whether models improve on the base rate and whether model upgrades help; Section~\ref{sec:category} examines category differences; and Section~\ref{sec:cutoff} studies the temporal patterns.

\section{Related Work}
\label{sec:related_work}

\paragraph{LLMs as probabilistic forecasters.}
Prior work evaluates LLMs as probabilistic forecasters under a range of information and aggregation settings.
\citet{halawi2024approaching} train retrieval-augmented systems that approach crowd-level calibration on questions from Metaculus, Good Judgment Open, INFER, Polymarket, and Manifold.
\citet{schoenegger2024wisdom} show that an ensemble of $12$ LLMs rivals a human crowd on $31$ binary Metaculus questions.
Without retrieval, single LLMs are near chance \citep{schoenegger2023large}; expert forecasters beat frontier LLMs on \textsc{ForecastBench} \citep{karger2025forecastbench, alur2025aia}; and frontier models are systematically overconfident on $300$ Kalshi questions, mostly worse than the base rate \citep{nel2025kalshi}.
Fine-tuning on Polymarket \citep{turtel2025llms} and scaling to frontier models \citep{lu2025evaluating} narrow but do not close the gap to human superforecasters.
Temporal information access remains a concern when interpreting forecasting performance \citep{paleka2025pitfalls, paleka2025consistency}.
LLMs memorize portions of their training data \citep{carlini2022quantifying}, and pre-/post-cutoff contamination detection is now standard \citep{golchin2024time, roberts2023data, shi2024detecting}.
\citet{lopezlira2025memorization} formalize this as a non-identification problem for economic forecasting, \citet{magar2022data} distinguish memorization from downstream exploitation, and \citet{li2026simulated} show that prompted ``forgetting'' of post-cutoff knowledge fails to replicate genuine temporal ignorance.
We examine forecasting performance jointly across training-cutoff strata, model variants, and event categories.

\paragraph{Prediction markets as ground truth.}
Prediction markets aggregate information through trader incentives \citep{wolfers2004prediction, berg2008long}, and trained forecasters outperform population baselines on such tasks \citep{mellers2014psychological, tetlock2005expert}; both motivate using crowd prices and human forecasters as benchmarks for LLMs.
Prior work compares against Metaculus participants \citep{schoenegger2024wisdom} or expert superforecasters \citep{karger2025forecastbench, mellers2015identifying}, both trained populations; Polymarket aggregates capital from a self-selected global pool of traders, a more representative deployment target.
Markets recover binary outcomes across domains \citep{dreber2015using} and often match or beat polls on identical questions \citep{atanasov2017distilling}, subject to crowd-wisdom conditions \citep{davisstober2014wise} and price-interpretation caveats under belief heterogeneity \citep{manski2006interpreting}.

\section{Setup}
\label{sec:setup}

\paragraph{\textsc{RMF}.}
\textsc{RMF} draws from the public Polymarket record of resolved binary markets between $2021$ and $2026$, spanning nine domains: crypto, entertainment, finance, geopolitics, politics, science/tech, sports, weather, and a residual \emph{other} category.
Polymarket resolutions are determined by financial settlement, which filters out ambiguous, unverifiable, or trivially-known questions and forces precise resolution criteria; the resulting ground truth is both large in scale and clean enough for stratified analysis at the domain level.

\paragraph{Models.}
We evaluate six models in three families, with two variants per family, so that each family supports a within-family scaling contrast (older$\to$newer for Claude, smaller$\to$larger for Qwen): Claude Sonnet (\textbf{4.5}, \textbf{4.6}), Claude Opus (\textbf{4.1}, \textbf{4.6}) \citep{anthropic2024claude}, and Qwen2.5 (\textbf{7B-Instruct} in bf16, \textbf{32B-Instruct-AWQ} in int4) \citep{qwen2024qwen25}.
The two Claude variants in each family share a tier (Sonnet at \$3 / \$15 per million input/output tokens, Opus at \$15 / \$75) but differ in training cutoff; Qwen variants share a cutoff (October~2023) but differ by roughly $4.5\times$ in parameter count.
Anthropic does not publicly disclose Claude parameter counts; full per-model release dates, vendor-published training cutoffs, and per-cutoff sample sizes are in App.~\ref{app:models}.

\begin{table}[h]
\centering
\small
\setlength{\tabcolsep}{4.5pt}
\caption{Models evaluated on the $n{=}3{,}000$ sample. ``n pre'' and ``n post'' use the paper's shared boundary (pre $\le$ 2024-12-31, post $\ge$ 2025-06-30). Cutoff dates for Sonnet~4.5 and Opus~4.1 are best-effort estimates from the corresponding release notes (Anthropic does not always publish an exact cutoff for intermediate model versions). For the Qwen2.5 checkpoints, both buckets fall after Qwen's own October~2023 cutoff and are therefore both strictly post-cutoff for those models; we still report counts under the paper's shared boundary so the rows are directly comparable.}
\label{tab:models}
\begin{tabular}{@{}llllrr@{}}
\toprule
\textbf{Model} & \textbf{API / HF ID} & \textbf{Released} & \textbf{Vendor cutoff} & \textbf{n pre} & \textbf{n post} \\
\midrule
Sonnet 4.5      & \texttt{claude-sonnet-4-5-20250929}    & 2025-09 & 2025-07           & $1{,}498$ & $1{,}502$ \\
Sonnet 4.6      & \texttt{claude-sonnet-4-6}             & 2025-12 & 2026-01           & $1{,}142$ & $1{,}152$ \\
Opus 4.1        & \texttt{claude-opus-4-1-20250805}      & 2025-08 & 2025-03           & $1{,}498$ & $1{,}502$ \\
Opus 4.6        & \texttt{claude-opus-4-6}               & 2025-11 & 2025-08           & $1{,}498$ & $1{,}501$ \\
Qwen2.5-7B      & \texttt{Qwen/Qwen2.5-7B-Instruct}      & 2024-09 & $\sim$2023-10     & $1{,}498$ & $1{,}502$ \\
Qwen2.5-32B-AWQ & \texttt{Qwen/Qwen2.5-32B-Instruct-AWQ} & 2024-09 & $\sim$2023-10     & $1{,}498$ & $1{,}502$ \\
\bottomrule
\end{tabular}
\end{table}

\paragraph{Training-data cutoffs.}
Vendor-published training cutoffs vary across the six models (App.~\ref{app:models}); the earliest is Qwen2.5 (October~2023) and the latest is Sonnet~4.6 (January~2026).
To keep \emph{post-cutoff} strictly out-of-distribution for all four Claude models simultaneously, we use a single conservative boundary calibrated to the earliest Claude cutoff.
A market is labeled \emph{pre-cutoff} if it resolved on or before December~31, 2024, \emph{post-cutoff} if it resolved on or after June~30, 2025, and excluded as ambiguous otherwise; the six-month exclusion zone serves as a buffer against soft-cutoff effects.
Both Qwen buckets fall strictly after Qwen's own October~2023 cutoff and so are post-cutoff for Qwen regardless of which Claude-calibrated bucket they land in (see App.~\ref{app:models}).

\paragraph{Protocol and metrics.}
Each model receives only the market question and outputs a probability $p \in [0,1]$ (zero-shot, no retrieval, no context).
We evaluate on the same $n{=}3{,}000$ stratified markets across all six models ($\sim$$1{,}500$ per period); per-model valid sample sizes range from $1{,}152$ to $1{,}502$ post-cutoff after parse failures (App.~\ref{app:models}).
We report the Brier score $\mathrm{BS} = \tfrac{1}{N} \sum_i (p_i - o_i)^2$ (lower is better) with bootstrap $95\%$ CIs ($10$K resamples), and the Murphy decomposition $\mathrm{BS} = \mathrm{REL} - \mathrm{RES} + \mathrm{UNC}$ \citep{murphy1973new}; full per-condition decomposition in App.~\ref{app:murphy}.
For pairwise comparison between two models we use the \emph{paired Brier difference} $\Delta_{A-B} = \tfrac{1}{|I|} \sum_{i \in I} \big[(p^A_i - o_i)^2 - (p^B_i - o_i)^2\big]$ over the set $I$ of markets where both models made a prediction, with a $95\%$ bootstrap CI over markets; a CI that contains zero indicates the two models are statistically indistinguishable on the matched set. Full procedure in App.~\ref{app:paired_brier}.

\paragraph{Baseline.}
We benchmark every model against the \emph{empirical base-rate predictor}, which assigns the marginal $P(\text{Yes})$ in each cutoff stratum as a constant prediction to every market in that stratum; we use the empirical rate rather than $0.5$ because Polymarket yes-rates are not balanced.
A model \emph{beats} the base rate when its Brier is strictly lower and its $95\%$ bootstrap CI excludes the baseline's Brier (the $^\ast$ marker in Table~\ref{tab:main}); we report the matched effect size as $\Delta_{\text{BR}} = \text{Brier}_{\text{base}} - \text{Brier}_{\text{model}}$.

\section{Results}
\label{sec:results}

We answer Q1 (do LLMs predict?) and Q2 (does scaling help?) in \S\ref{sec:capacity}, identify the dominant axis (event category) in \S\ref{sec:category}, and characterize the cutoff as a moderator in \S\ref{sec:cutoff}.

\subsection{Frontier prediction barely beats base rate; scaling helps only below the frontier}
\label{sec:capacity}

\paragraph{Q1: Can LLMs predict the future well? Barely.}
All four Claude models converge to a narrow Brier band around $0.19$ (Table~\ref{tab:main}), beating the base rate by only $0.024$--$0.033$: significant at $n{>}1{,}000$ per stratum, practically marginal.
The two Qwen models trail: 32B is indistinguishable from the base rate, and 7B is significantly \emph{worse} than it.

\begin{table}[t!]
\centering
\small
\caption{Post-cutoff Brier across all six models. $\Delta_{\text{BR}}$: base-rate Brier $-$ model Brier (positive $=$ model beats base rate). $^\ast$ $=$ base rate outside the model's $95\%$ CI. All four Claude models clear the base rate by $0.02$--$0.03$ Brier; Qwen 32B is indistinguishable from it; Qwen 7B is significantly worse.}
\label{tab:main}
\begin{tabular}{lrccc}
\toprule
\textbf{Model} & $n$ & \textbf{Brier} & \textbf{Base rate} & \textbf{$\Delta_{\text{BR}}$} \\
\midrule
Sonnet 4.5 & $1{,}502$ & $.186$ & $.216$ & $+.030^\ast$ \\
Sonnet 4.6 & $1{,}152$ & $.189$ & $.217$ & $+.028^\ast$ \\
Opus 4.1   & $1{,}502$ & $.183$ & $.216$ & $+.033^\ast$ \\
Opus 4.6   & $1{,}501$ & $.192$ & $.216$ & $+.024^\ast$ \\
Qwen 7B    & $1{,}502$ & $.234$ & $.216$ & $-.018^\ast$ \\
Qwen 32B   & $1{,}502$ & $.210$ & $.216$ & $+.006$ \\
\bottomrule
\end{tabular}
\end{table}

\paragraph{Q2: Does scaling capacity help? Only below the frontier.}
Table~\ref{tab:cap_pairs} reports five paired contrasts: three within-family scaling tests (Sonnet 4.5$\to$4.6, Opus 4.1$\to$4.6, Qwen 7B$\to$32B) and two cross-family tests (Sonnet vs.\ Opus within each generation).
Only one is significant in the expected direction: Qwen 7B$\to$32B.
Inside the Claude lineup, none of the four contrasts improves significantly, and Opus 4.1$\to$4.6 is significantly \emph{negative}.
Capacity helps when a model is weak (lifting Qwen from below the base rate to indistinguishable from it), but once a model reaches the Claude convergence band near $0.19$, neither version nor tier moves the needle.
The Opus pair has few markets in the memorization-gap window (App.~\ref{app:cutoff_pairs}), so its negative paired difference reflects post-cutoff prediction, not memorization.

\begin{table}[t!]
\centering
\small
\caption{Paired Brier difference $\Delta$ (later minus earlier; negative $=$ later is better) on post-cutoff markets. $^\ast$ $=$ 95\% bootstrap CI excludes zero. Only Qwen 7B$\to$32B improves significantly; within-Claude contrasts are null or significantly negative. Full CIs in App.~\ref{app:paired_brier}.}
\label{tab:cap_pairs}
\begin{tabular}{@{}lc@{}}
\toprule
\textbf{Contrast} & \textbf{$\Delta$} \\
\midrule
\multicolumn{2}{@{}l}{\textit{Within-family version (older $\to$ newer)}} \\
\quad Sonnet 4.5 $\to$ 4.6     & $+.004$ \\
\quad Opus 4.1 $\to$ 4.6       & $+.009^\ast$ \\
\multicolumn{2}{@{}l}{\textit{Within-family size (smaller $\to$ larger)}} \\
\quad Qwen 7B $\to$ 32B        & $-.024^\ast$ \\
\multicolumn{2}{@{}l}{\textit{Cross-family capacity (Sonnet $\to$ Opus)}} \\
\quad Within 4.6 generation    & $+.004$ \\
\quad Within 4.5/4.1 generation & $-.003$ \\
\bottomrule
\end{tabular}
\end{table}

\subsection{Category: matters far more than capacity}
\label{sec:category}

Event category affects post-cutoff Brier substantially more than capacity does within the frontier.
Within-model Brier spreads across categories are several times larger than the largest within-Claude paired contrast (\S\ref{sec:capacity}) and the Qwen size effect (Table~\ref{tab:cat_spread}; full per-cell numbers in App.~\ref{app:cat_breakdown}).
Best and worst categories also differ across models, so no universal LLM-friendly domain exists.

\begin{table}[t!]
\centering
\small
\caption{Within-model post-cutoff Brier spread across categories ($n{\ge}20$ per cell). The spread is several times the largest paired capacity effect from \S\ref{sec:capacity}; best and worst categories differ across models. Full per-cell numbers in App.~\ref{app:cat_breakdown}.}
\label{tab:cat_spread}
\setlength{\tabcolsep}{4pt}
\begin{tabular}{@{}lcccc@{}}
\toprule
\textbf{Model} & \textbf{Best cat} & \textbf{Brier} & \textbf{Worst cat} & \textbf{Brier} \\
\midrule
Sonnet 4.5 & weather     & $.126$ & sports     & $.195$ \\
Sonnet 4.6 & geopolitics & $.138$ & sports     & $.205$ \\
Opus 4.1   & geopolitics & $.102$ & sports     & $.205$ \\
Opus 4.6   & weather     & $.129$ & geopolitics & $.229$ \\
Qwen 7B    & geopolitics & $.149$ & sci/tech   & $.274$ \\
Qwen 32B   & geopolitics & $.143$ & other      & $.225$ \\
\bottomrule
\end{tabular}
\end{table}

\begin{table}[h]
\centering
\small
\caption{Post-cutoff Brier by event category and model. Per-category $n$ in parentheses (varies slightly for Sonnet 4.6 because of parse failures). Italic cells ($n<20$) are directional only.}
\label{tab:cat_breakdown}
\begin{tabular}{@{}lcccccc@{}}
\toprule
\textbf{Category} & \textbf{Sonnet 4.5} & \textbf{Sonnet 4.6} & \textbf{Opus 4.1} & \textbf{Opus 4.6} & \textbf{Qwen 7B} & \textbf{Qwen 32B} \\
\midrule
crypto        & $.176$ ({\small 178}) & $.170$ ({\small 136}) & $.164$ ({\small 178}) & $.165$ ({\small 177}) & $.203$ ({\small 178}) & $.192$ ({\small 178}) \\
entertainment & \textit{$.150$ ({\small 14})} & \textit{$.036$ ({\small 10})} & \textit{$.150$ ({\small 14})} & \textit{$.160$ ({\small 14})} & \textit{$.199$ ({\small 14})} & \textit{$.219$ ({\small 14})} \\
finance       & \textit{$.197$ ({\small 14})} & \textit{$.187$ ({\small 11})} & \textit{$.101$ ({\small 14})} & \textit{$.218$ ({\small 14})} & \textit{$.193$ ({\small 14})} & \textit{$.202$ ({\small 14})} \\
geopolitics   & $.147$ ({\small 34}) & $.138$ ({\small 28}) & $.102$ ({\small 34}) & $.229$ ({\small 34}) & $.149$ ({\small 34}) & $.143$ ({\small 34}) \\
other         & $.198$ ({\small 745}) & $.200$ ({\small 566}) & $.198$ ({\small 745}) & $.202$ ({\small 745}) & $.248$ ({\small 745}) & $.225$ ({\small 745}) \\
politics      & $.183$ ({\small 64}) & $.165$ ({\small 48}) & $.179$ ({\small 64}) & $.219$ ({\small 64}) & $.220$ ({\small 64}) & $.205$ ({\small 64}) \\
science/tech  & $.178$ ({\small 95}) & $.190$ ({\small 71}) & $.159$ ({\small 95}) & $.170$ ({\small 95}) & $.274$ ({\small 95}) & $.179$ ({\small 95}) \\
sports        & $.195$ ({\small 243}) & $.205$ ({\small 191}) & $.205$ ({\small 243}) & $.208$ ({\small 243}) & $.229$ ({\small 243}) & $.216$ ({\small 243}) \\
weather       & $.126$ ({\small 115}) & $.166$ ({\small 91}) & $.123$ ({\small 115}) & $.129$ ({\small 115}) & $.210$ ({\small 115}) & $.171$ ({\small 115}) \\
\bottomrule
\end{tabular}
\end{table}

\subsection{Cutoff: a moderator of category, not a main effect}
\label{sec:cutoff}

If models genuinely forecasted, the cutoff axis would be irrelevant.
If models purely recalled, crossing the cutoff would cause a uniform Brier collapse.
We find neither: pre- and post-cutoff Brier are within $0.03$ for every Claude model, with no sharp memorization cliff as the Opus buffer varies from $0$ to $360$ days (App.~\ref{app:marginals}, App.~\ref{app:temporal}).
As a main effect, the cutoff buys almost nothing.

\begin{table}[t!]
\centering
\small
\caption{Brier marginals across all six models, both periods. $\Delta_{\text{BR}}$: base-rate minus model Brier on post-cutoff (positive = model beats base rate). $^\ast$ = base rate outside the model's post-cutoff $95\%$ CI.}
\label{tab:marginals}
\begin{tabular}{lcccc}
\toprule
\textbf{Model} & \textbf{n post} & \textbf{Pre} & \textbf{Post} & \textbf{$\Delta_{\text{BR}}$} \\
\midrule
Sonnet 4.5 & $1{,}502$ & $.196$ {\scriptsize $[.184,.208]$} & $.186$ {\scriptsize $[.176,.196]$} & $+.030^\ast$ \\
Sonnet 4.6 & $1{,}152$ & $.220$ {\scriptsize $[.205,.236]$} & $.189$ {\scriptsize $[.178,.201]$} & $+.028^\ast$ \\
Opus 4.1   & $1{,}502$ & $.211$ {\scriptsize $[.198,.224]$} & $.183$ {\scriptsize $[.173,.193]$} & $+.033^\ast$ \\
Opus 4.6   & $1{,}501$ & $.211$ {\scriptsize $[.198,.225]$} & $.192$ {\scriptsize $[.181,.203]$} & $+.024^\ast$ \\
Qwen 7B    & $1{,}502$ & $.264$ {\scriptsize $[.251,.277]$} & $.234$ {\scriptsize $[.222,.246]$} & $-.018^\ast$ \\
Qwen 32B   & $1{,}502$ & $.231$ {\scriptsize $[.220,.243]$} & $.210$ {\scriptsize $[.200,.220]$} & $+.006$ \\
\bottomrule
\end{tabular}
\end{table}

The cutoff signal does not disappear; it concentrates inside the category axis.
Stratifying each model's markets into log-volume tertiles as a proxy for salience (App.~\ref{app:salience}), the salience gap turns from non-positive pre-cutoff to positive post-cutoff for every Claude model and for Qwen 32B: prominent events are easy when memorizable and hard when novel.
A second signature is in cross-model agreement: mean per-event squared-error correlation rises from $0.42$ pre-cutoff to $0.62$ post-cutoff, with the lift concentrated inside the frontier Claude lineup (App.~\ref{app:cross_model}); the frontier converges on the same hedged predictions.
The cutoff matters precisely where memorization could matter, and nowhere else.

\section{Conclusion}
\label{sec:conclusion}

We introduce \textbf{RMF}, a benchmark for evaluating LLM probability forecasts, using each model's training cutoff to separate real prediction from recall.
Across six models in three families, three findings emerge: frontier LLMs barely beat the empirical base-rate predictor; capacity scaling helps only below the frontier and stops once a model reaches the Claude convergence band; and event category drives more variance in post-cutoff Brier than capacity does.
Better reasoning benchmarks do not produce better frontier forecasters.

\section*{Limitations}

Three limitations.
\emph{(i) No ceiling baseline.} We benchmark only against the empirical base rate, not against Polymarket crowd prices or expert forecasters; the gap to what is achievable is unknown.
\emph{(ii) Zero-shot, no-retrieval protocol.} Predictions use internalized knowledge alone; allowing internet access would likely shrink the gap to the base rate and may change the within-frontier scaling story.
\emph{(iii) Time-since-cutoff decay.} We treat all post-cutoff markets as equivalent, but markets resolving weeks vs.\ months past a cutoff differ; we do not decompose Brier by time-since-cutoff.

\section*{Ethical Considerations}
Probabilistic LLM forecasters carry a deployment risk: users tend to trust calibrated-looking numerical outputs more than the underlying reliability warrants, especially in financial and policy settings where a confident probability can drive concrete decisions.
We release RMF so that deployers can directly measure per-model reliability against ground truth instead of relying on aggregate benchmark scores or vendor claims.
On the data side, RMF uses only public post-resolution Polymarket metadata (question, outcome, resolution date, category); we access no trader identities, order-book data, or live prices, and contribute no order flow to Polymarket itself.
Frontier model versions are deprecated by vendors over time; we list exact model identifiers and vendor-published training cutoffs in App.~\ref{app:models} to support replication while the evaluated versions remain available.

\section*{Acknowledgements}
We used Anthropic's Claude Opus 4.7 for writing assistance and prose editing on this manuscript; all research design, experiments, analyses, and findings are the authors'.

\bibliography{references}
\bibliographystyle{colm2026_conference}

\appendix

\section{Models Evaluated on RMF}
\label{app:models}

Table~\ref{tab:models} lists every model evaluated in this study, with vendor-published training cutoffs, release dates, and per-cutoff $n{=}3{,}000$ sample sizes.
Each family contains two variants, supporting a within-family scaling contrast: older$\to$newer for the two Claude families (Sonnet, Opus) and smaller$\to$larger for Qwen2.5.
The Qwen2.5-7B-Instruct checkpoint runs in bf16; the 32B-Instruct-AWQ checkpoint is quantized to int4 via AWQ for single consumer GPU inference.

\section{Pairwise Cutoff Coverage for Version-Robustness Pairs}
\label{app:cutoff_pairs}

The version-robustness checks in \S\ref{sec:capacity} compare two pairs of Claude models that share an architecture family but differ in training-data cutoff: Sonnet~4.5 (2025-07) vs.\ Sonnet~4.6 (2026-01), and Opus~4.1 (2025-03) vs.\ Opus~4.6 (2025-08).
Table~\ref{tab:cutoff_pairs} splits the $n{=}3{,}000$ sample by where each market's resolution date falls relative to both members of a pair:
(i) \emph{both pre}, in training for both members and so a paired retrieval comparison;
(ii) \emph{split}, in training for the newer member only, the window in which any memorization-gap effect should appear; and
(iii) \emph{both post}, out of training for both members and therefore a paired comparison on genuine prediction.
Counts are over the markets where both members returned a parseable prediction (the intersection on which paired Brier contrasts can be computed).

\begin{table}[h]
\centering
\small
\caption{Pairwise cutoff coverage on the $n{=}3{,}000$ sample, restricted to markets jointly parseable by both models in the pair. ``Split'' is the memorization-gap window: only the newer member could plausibly have ingested these outcomes during training. The Opus pair has very few split markets ($n{=}51$) because the $n{=}3{,}000$ sample was constructed under the shared 2024-12-31 / 2025-06-30 boundary and the Opus~4.1 to Opus~4.6 gap (2025-03 to 2025-08) falls almost entirely inside the excluded buffer; consequently, the Opus pair primarily contrasts both-pre and both-post performance, not the memorization-gap window.}
\label{tab:cutoff_pairs}
\begin{tabular}{@{}lrrrr@{}}
\toprule
\textbf{Pair (older vs.\ newer)} & \textbf{Both pre} & \textbf{Split} & \textbf{Both post} & \textbf{Total} \\
\midrule
Sonnet 4.5 (cutoff 2025-07) vs.\ Sonnet 4.6 (cutoff 2026-01) & $1{,}150$ & $480$ & $664$    & $2{,}294$ \\
Opus 4.1 (cutoff 2025-03) vs.\ Opus 4.6 (cutoff 2025-08)     & $1{,}498$ & $51$  & $1{,}450$ & $2{,}999$ \\
\bottomrule
\end{tabular}
\end{table}

\section{Murphy Brier Decomposition}
\label{app:murphy}

Table~\ref{tab:murphy} presents the full Murphy decomposition for Opus predictions.
The Brier score of a single model decomposes as $\text{BS} = \text{REL} - \text{RES} + \text{UNC}$, where reliability (REL; lower is better) measures calibration error, resolution (RES; higher is better) measures discriminative ability, and uncertainty (UNC) is the irreducible base-rate variance.

\begin{table}[t!]
\centering
\small
\caption{Murphy decomposition of Opus Brier score on the $n{=}3{,}000$ sample. Resolution drops on novel events (less discrimination) but reliability improves (better calibration).}
\label{tab:murphy}
\begin{tabular}{lccc}
\toprule
\textbf{Component} & \textbf{Pre} & \textbf{Post} & \textbf{$\Delta$} \\
\midrule
Reliability $\downarrow$ & $.023$ {\scriptsize $[.016,.034]$} & $.012$ {\scriptsize $[.007,.021]$} & $-.011$ \\
Resolution $\uparrow$  & $.048$ {\scriptsize $[.040,.061]$} & $.034$ {\scriptsize $[.027,.045]$} & $-.014$ \\
Uncertainty            & $.233$ & $.215$ & $-.018$ \\
\midrule
\textbf{Brier} $\downarrow$ & $.207$ & $.193$ & $-.014$ \\
\bottomrule
\end{tabular}
\end{table}

\section{Paired Brier Difference}
\label{app:paired_brier}

For two models $A$ and $B$, the paired Brier difference on the intersection $I$ of markets where both made a prediction is

$$\Delta_{A-B} = \frac{1}{|I|} \sum_{i \in I} \left[ (p^A_i - o_i)^2 - (p^B_i - o_i)^2 \right]$$

We report a $95\%$ bootstrap CI by resampling markets in $I$ with replacement ($10$K resamples) and taking the $2.5$th and $97.5$th percentiles of the recomputed $\Delta$.
We use the paired difference rather than the difference of aggregate Briers because pairing removes the variance contributed by question difficulty: even when two models share an aggregate Brier they may disagree on which questions are easy, and the paired statistic isolates the model-driven contribution from that question-driven contribution.
We restrict to the intersection rather than imputing missing predictions because parse failures are not random across models, and treating them as either zero, the model's mean prediction, or a uniform $0.5$ would each bias $\Delta$ in a different direction.
The sign convention used throughout the paper is $A - B$ for ``$A$ minus $B$'' contrasts (so $\Delta < 0$ when $B$ has lower mean squared error and is therefore the better predictor).
Table~\ref{tab:paired_ci} reports the full $95\%$ CI for every paired contrast in Table~\ref{tab:cap_pairs}.

\begin{table}[t!]
\centering
\small
\caption{Paired Brier difference with $95\%$ bootstrap CI for the five contrasts in Table~\ref{tab:cap_pairs}. $n$ is the intersection size.}
\label{tab:paired_ci}
\begin{tabular}{@{}lrcl@{}}
\toprule
\textbf{Contrast} & $n$ & \textbf{$\Delta$} & \textbf{[95\% CI]} \\
\midrule
Sonnet 4.5 $\to$ 4.6     & $1{,}152$ & $+.004$       & $[-.004, +.011]$ \\
Opus 4.1 $\to$ 4.6       & $1{,}501$ & $+.009^\ast$  & $[+.001, +.017]$ \\
Qwen 7B $\to$ 32B        & $1{,}502$ & $-.024^\ast$  & $[-.035, -.014]$ \\
Sonnet 4.6 vs.\ Opus 4.6 & $1{,}152$ & $+.004$       & $[-.005, +.014]$ \\
Sonnet 4.5 vs.\ Opus 4.1 & $1{,}502$ & $-.003$       & $[-.011, +.004]$ \\
\bottomrule
\end{tabular}
\end{table}

\section{Post-Cutoff Brier by Category and Model}
\label{app:cat_breakdown}

Table~\ref{tab:cat_breakdown} reports post-cutoff Brier by event category for all six models on the $n{=}3{,}000$ sample.
Per-cell $n$ is identical across models within a category (the sample is shared), with small variations from parse failures on Sonnet 4.6.
Cells with $n<20$ (entertainment, finance) are shown in italics as directional only.
Within a single column (a model), the spread across categories is several times the spread across columns within a single row (a category), supporting the category-over-capacity claim in \S\ref{sec:category}.

\section{Temporal Ablation}
\label{app:temporal}

To test whether performance depends on the exact cutoff buffer, we sweep the buffer width from 0 to 360~days.
Post-cutoff Brier remains in $[0.180, 0.195]$ across all buffer widths, and the pre/post gap stays small and negative throughout.
There is no sharp ``memorization cliff''---consistent with a gradual rather than step-function boundary between memorized and novel events.

\section{Pre/Post Brier Marginals Across All Six Models}
\label{app:marginals}

Table~\ref{tab:marginals} reports pre- and post-cutoff Brier with bootstrap $95\%$ CIs (10K resamples) and the post-cutoff base-rate predictor for each model.
All four Claude models significantly beat the base rate post-cutoff; Qwen 32B is indistinguishable from it; Qwen 7B is significantly worse.

\section{Salience-Tertile Breakdown}
\label{app:salience}

We split each model's markets into log-volume tertiles (low / mid / high) and recompute Brier within each cutoff condition (Table~\ref{tab:salience}).
The salience gap (high $-$ low Brier) is non-positive pre-cutoff for all four Claude models and turns positive post-cutoff for each, so $\Delta_\text{gap}$ isolates a memorization-on-prominent-events effect: prominent markets are relatively easy when memorizable and relatively hard when novel.
Qwen 32B shows the same pattern; Qwen 7B does not ($\Delta_\text{gap}{=}-0.022$), consistent with 7B being too weak to memorize even prominent events.

\begin{table}[t!]
\centering
\small
\caption{Brier by log-volume salience tertile, cutoff, and model.}
\label{tab:salience}
\begin{tabular}{lcccc}
\toprule
& \textbf{low} & \textbf{mid} & \textbf{high} & \textbf{gap} \\
\midrule
\multicolumn{5}{l}{\textit{Sonnet 4.5}} \\
pre  & $.217$ & $.176$ & $.190$ & $-.026$ \\
post & $.195$ & $.177$ & $.214$ & $+.019$ \\
\multicolumn{4}{l}{$\Delta_\text{gap}$} & $+.045$ \\
\midrule
\multicolumn{5}{l}{\textit{Sonnet 4.6}} \\
pre  & $.221$ & $.226$ & $.216$ & $-.005$ \\
post & $.202$ & $.171$ & $.220$ & $+.018$ \\
\multicolumn{4}{l}{$\Delta_\text{gap}$} & $+.023$ \\
\midrule
\multicolumn{5}{l}{\textit{Opus 4.1}} \\
pre  & $.220$ & $.199$ & $.214$ & $-.006$ \\
post & $.196$ & $.168$ & $.209$ & $+.013$ \\
\multicolumn{4}{l}{$\Delta_\text{gap}$} & $+.019$ \\
\midrule
\multicolumn{5}{l}{\textit{Opus 4.6}} \\
pre  & $.222$ & $.206$ & $.209$ & $-.014$ \\
post & $.207$ & $.182$ & $.213$ & $+.005$ \\
\multicolumn{4}{l}{$\Delta_\text{gap}$} & $+.019$ \\
\midrule
\multicolumn{5}{l}{\textit{Qwen 7B}} \\
pre  & $.251$ & $.261$ & $.273$ & $+.022$ \\
post & $.243$ & $.222$ & $.244$ & $+.000$ \\
\multicolumn{4}{l}{$\Delta_\text{gap}$} & $-.022$ \\
\midrule
\multicolumn{5}{l}{\textit{Qwen 32B}} \\
pre  & $.247$ & $.217$ & $.229$ & $-.018$ \\
post & $.215$ & $.207$ & $.228$ & $+.014$ \\
\multicolumn{4}{l}{$\Delta_\text{gap}$} & $+.032$ \\
\bottomrule
\end{tabular}
\end{table}

\section{Per-Event Cross-Model Agreement}
\label{app:cross_model}

Restricting to the intersection where all six models made a prediction ($n{=}1{,}142$ pre-cutoff, $n{=}1{,}152$ post-cutoff), Table~\ref{tab:cross} reports the mean per-event squared-error correlation across pairs of models, by pair group.
The overall mean rises from $0.42$ pre-cutoff to $0.62$ post-cutoff: novel-event errors are shared across models, consistent with a shared base-rate hedging strategy.
The lift is largest within the frontier Claude lineup (mean $0.53 \to 0.75$); the four Claude models converge tightly on the same predictions for events none of them have seen.
Within Qwen the lift is much smaller ($0.51 \to 0.56$), and cross-family Claude$\times$Qwen pairs lie in between ($0.32 \to 0.53$).

\begin{table}[t!]
\centering
\small
\caption{Mean per-event squared-error correlation by pair group, on the all-six-model intersection. Within-Claude correlations rise sharply post-cutoff; within-Qwen does not.}
\label{tab:cross}
\begin{tabular}{lccc}
\toprule
\textbf{Group} & \textbf{$n$ pairs} & \textbf{Pre} & \textbf{Post} \\
\midrule
Within Claude (4 models)         & $6$  & $.53$ & $.75$ \\
Within Qwen (2 models)           & $1$  & $.51$ & $.56$ \\
Cross-family (Claude $\times$ Qwen) & $8$  & $.32$ & $.53$ \\
\midrule
Overall                           & $15$ & $.42$ & $.62$ \\
\bottomrule
\end{tabular}
\end{table}

\section{Artifact Licenses}
\label{app:licenses}

We will release the \textsc{RMF} data under CC-BY 4.0 and the evaluation code under the MIT License.
The LLMs we evaluate are used under their respective vendor terms: Claude models via the Anthropic API under Anthropic's usage policies \citep{anthropic2024claude}; Qwen2.5 7B-Instruct and 32B-Instruct-AWQ under the licenses distributed with their respective Hugging Face releases \citep{qwen2024qwen25}.
The underlying Polymarket data is public post-resolution market metadata; we accessed no non-public data and contributed no order flow to the platform.
All uses are consistent with the artifacts' intended purposes: the model APIs and open-weight releases are distributed for research and downstream evaluation, the Polymarket record is public historical market metadata, and \textsc{RMF} is released for non-commercial research use in LLM forecasting evaluation.

\end{document}